\documentclass{article}
\usepackage{ijcai20}

\usepackage{times}
\usepackage{soul}
\usepackage{url}
\usepackage[hidelinks]{hyperref}
\usepackage[utf8]{inputenc}
\usepackage[small]{caption}
\usepackage{graphicx}
\usepackage{amsmath}
\usepackage{amsthm}
\usepackage{booktabs}
\usepackage{algorithm}
\usepackage{algorithmic}
\usepackage{comment}

\usepackage{lscape}
\usepackage{natbib}
\usepackage{graphicx}
\usepackage[table]{xcolor}
\definecolor{lightgray}{gray}{0.9}
\usepackage{subfig}
\usepackage{fullpage}
\usepackage{times}
\usepackage{fancyhdr,graphicx,amsmath,amssymb}

\newcommand{\circled}[1]{(#1)}

\title{Heterogeneous Data-Aware Federated Learning}

\author{
Lixuan YANG
\and
Cedric BELIARD\and
Dario ROSSI
\affiliations
Huawei Technologies Co. Ltd 
\emails
\{lixuan.yang, cedric.beliard, dario.rossi\}@huawei.com,

}

\begin{document}

\maketitle

\begin{abstract}
Federated learning (FL) is an appealing concept to perform distributed training of Neural Networks (NN)  while keeping data private.  With the industrialization of the FL framework, we identify several problems hampering its successful deployment, such as presence of non i.i.d data, disjoint classes, signal multi-modality across datasets. In this work, we address these problems by proposing a novel method that not only  (1) aggregates generic model parameters (e.g. a common set of task generic NN layers) on server (e.g. in traditional FL), but also (2) keeps a set of  parameters (e.g, a set of task specific NN layer) specific to each client. 
We validate our method on the traditionally used public benchmarks (e.g., Femnist) as well as on our proprietary collected dataset (i.e., traffic classification). Results show the benefit of our method, with significant advantage on extreme cases.

\end{abstract}

\section{Introduction}

\rowcolors{3}{}{lightgray}
\begin{table*}[!th]
  \begin{center}
  \begin{small}
    \caption{Summary of the State of the Art Federated Learning Methods}
    \label{tab:related}
    \begin{tabular}{lccccc}
      \textbf{Paper} & \textbf{Client} & \textbf{Aggregation} & \textbf{Loss} & \textbf{Information} & \textbf{Model}\\
        \textbf{[Year]}&   \textbf{Selection} & \textbf{function}& \textbf{function} &  \textbf{Exchanged} & \\
      \hline
      FedAvg \cite{McMahanMRA16} & Random & Avg & - & Model & - \\
      FedProx \cite{Kumar2018} & Random & Avg & Custom & Model & - \\ 
      \cite{Zhao2018} & Random & Avg & - & Model+Data(S2C) & - \\
      \cite{Anil2018} & Random & - & Distillation & Model & -\\
      \cite{Jeong2018} & All & Avg & Distillation & Each class 1 vector$\star$ & - \\
      Agnos \cite{Mohri2019} & - & Avg & Custom & Distribution Update & -\\
      LG \cite{Liang2019} & Random & Avg & \shortstack{Softmax+\\feature learning} & partial model & \shortstack{ Feature Extractor + \\Class Predictor}\\
      \cite{Sattler2019} & Clustering & Avg & - & Model & - \\
      q-FFL\cite{Tian2019} & Random & Reweight Avg & - & Model+qValue & - \\
      \cite{Izbicki2019} & - & Reweight Avg & - & Model & - \\
            \hline
    \emph{this paper}  & \emph{Random} & \emph{Avg} & - & \emph{generic parameters} & \emph{generic + specific}  \\
    \end{tabular}
  \end{small}
  \end{center}
\end{table*}

Machine learning and particularly Neural Network (NN) models are now tremendously popular to solve computer vision, natural language processing and networking problems. Proper \emph{model training in centralized settings} is not without challenges. On the one hand, centralized training raises data sharing concerns, both in terms of transfer volumes and data privacy -- which in European countries is very critical due to the recent stringent regulation on private data protection (GDPR).
On the other hand, model training on individual datasets helps solving data sharing concerns, yet generates portability/generalization issues.
To circumvent the above concerns, \cite{McMahanMRA16}  introduced the notion of \emph{Federated Learning (FL)}, that has a number of appealing properties. First and foremost, FL allows for collaboratively training the model, without having to share the data. Second, sharing models allow additionally to save significant data transfer volume, yielding beneficial cost reduction in cloud-based models (as charging also depends on data transfer). Third, distributed  training also reinforces the generalization capabilities of the model.

However, FL is not without downsides. 
First and foremost,  even \circled{0} sharing models can be a sensitive issue, either due to the presence of multiple stackholders, or due to stringent regulation. Additionally, in practical deployment scenarios data exhibit properties  such as \circled{1} Class imbalance, \circled{2} Disjoint class distribution and \circled{3} Signal multi-modality within each class across datasets,
which lead to model divergence and  could limit the practical relevance of  FL models.
To make FL viable in presence of \circled{0} model sharing constraints, while at the same time handling the above \circled{1}-\circled{3} data characteristics, we propose in this paper a novel method where clients are not only able to benefit from  model sharing, but also and especially retain the ability to \emph{keep secret} and \emph{specialize} part of their model, to respectively handle problems  \circled{0}  and \circled{1}-\circled{3}. 
We argue this can be achieved by \emph{divide et impera}: in a nutshell, in the NN architecture we isolate a common generic parameters from the client-specific parameters, and treat them differently in the distributed training process.
In particular, we propose that each client keeps a private and independent classifier, while sharing the common feature extraction process by averaging the generic parameters: so doing, clients benefit by the global learning process, whilst being able to keep their own (private) differences at the same time. 

We perform a thorough performance evaluation of our proposed Heterogeneous Data Aware Federated Learning (HDAFL) approach, contrasting it with the classic benchmark in the state of the art, namely~\cite{McMahanMRA16}. We leverage different datasets, including the classic FEMNIST dataset, which exhibits characteristics (1)-(2), as well as very large proprietary dataset for TCP/IP traffic classification which which exhibits characteristics (1)-(3). Results show that  HDAFL  improves the local and global system accuracy, even slightly reducing the volume of information exchanged, which makes FL of practical relevance in real-world situations.

In the rest of this paper, we overview the related work (Sec 2), and contextualize our applications and dataset (Sec 3). We next outline our  method (Sec 4),  contrast its performance with the state of the art (Sec 5) and  summarize our findings (Sec 6).

\section{State of the art}
\cite{McMahanMRA16}  pioneered FL by proposing the Federated Averaging (FedAvg) algorithm. This 
 ignited  valuable FL research,  that we compactly summarize in  Tab.\ref{tab:related} and briefly overview in this section.

\subsection{Complexity}
A first wave of research was directed into reducing the communication complexity. For instance, 
\cite{Konecn2016} proposed the model compression and random sampling to send less data possible in each round of communication. \cite{Izbicki2019} proposed an extreme case of one round of communication that used a subset of data in server to estimate the optimal merge weights subspace. Other proposals use knowledge distillation for the model aggregation as in \cite{Anil2018}, that propose a CoDistillation that the clients use the other clients' model as teacher model. To economise the communication cost, In \cite{Jeong2018}, the clients uses directly the average predictions of each class as teacher's output, these predictions are aggregated in server.

Another stringent constraint in real world is 
the computational bottleneck due to system heterogeneity: \cite{Kumar2018} propose a $\gamma$-inexactness to measure the amount of local computation from the local solver, while \cite{Tian2019} assign higher weights to devices with poor performance in order to reduce variance.

\subsection{Data characteristics}
While the above methods works well on ideal data, ~\cite{Sattler2019, Zhao2018} proved that FedAvg diverges in presence of non i.i.d. data or adversarial behavior:  this fueled further research to tackle specific aspects of the data, closer to our work.
For instance, to counter the multi-modality problem, the Clustered federated learning proposed by \cite{Sattler2019} groups clients by using the cosine similarity of gradient updates, and applying FedAvg in each clusters (for clusters that fail to converge, the clients in that cluster will be bi-partitioned).

For non i.i.d data, not only the global distribution is unknown as pointed out by \cite{Zhao2018}, but and the model divergence coming from weights difference is fundamentally rooted in the data distribution distance
as proven in \cite{Mohri2019}.  To counter these problems, \cite{Zhao2018} finds that the globally shared data (from server back to clients) reduces weight divergence, which however poses additional privacy problems.  \cite{Mohri2019} instead propose  to optimize a global loss function over all possible distributions. Finally, \cite{Kumar2018} handles non i.i.d data by restricting local updates to be closer to the global model.

Closest to our work is Local Global FL (LG-FedAvg) by~\cite{Liang2019}, which relies on local data to learn representation on devices and distribute a partial  model that predicts on compact feature. Interestingly, LG-FedAvg is similar to our work, as \emph{only part} of the model is shared: however, we make an opposite choice wrt LG-FedAvg about \emph{which part} of the model should be shared, as LG-FedAvg aggregate the specific parameters in the central server. 
Roughly speaking, the success of CNN is mainly based on the feature extraction, the specific parameters are just a projection of the feature vector: thus, LG-FedAvg supposes that each client has enough data to train a local feature extractor, which we instead argue not to be always the case. In contrast with LG-FedAvg, we aggregate the common generic parameters of initial layers, i.e., the process of feature extraction. Consequently, client profits from the rich feature extraction from entire data, while further being able to specialize the specific layer to comply with  to cope with heterogeneous data, unlike LG-FedAvg.

\section{Applications (and Datasets)}
\label{sec:dataset}
For a stackholder such as Huawei, the field of application for AI technologies is broader than the classic image recognition domain, and includes workflows typical of Internet Service Providers. As such, we consider here two example of applications, notably (1) image recognition and (2) the classification of Internet traffic, along with two respective datasets, whose volumes are compactly summarized and illustrated in Fig.~\ref{fig:dataset}. In this section, we describe relevant characteristics of the datasets, further examplifying the aforementioned properties.

\subsection{Image recognition (FEMNIST)}
As typically done in the FL literature, we consider an image recognition application. In the LEAF benchmark, we selected the Federated Extended Mnist (FEMNIST) dataset. The FEMNIST dataset  comprises 805k  samples of 62 classes including digits and character, and is extensively described by~\cite{Cohen17, Caldas18}, to which we refer the interested reader. 
As this dataset  is the most commonly used in the FL literature \cite{Tian2019}, this allows to readily put results in perspective with the related work.


\subsection{TCP/IP Traffic classification  (Huawei)}
We leverage a large proprietary dataset,  collected at two different clients representative of Home and Enterprise business respectively, for the purpose of network traffic classification, which is a key function of network traffic management. 
Class labels represent fine-grained Internet applications, while input signal are  timeseries of the size of the first 10 packets, which are known to have a high discriminative power for the classification task by ~\cite{ccr07} and be amenable to real-time inference by~\cite{imc12}, and are well suited as input of a 1D-CNN architecture.  This dataset represent a different operational point as it has more than twice of the samples and about half of the classes than FEMNIST.

\begin{table}[!t]
  \begin{center}
    \caption{Datasets at a glance.} \label{tab:dataset}
    \begin{tabular}{lcc}
     {\bf Application} & \shortstack{{\bf Image}\\{\bf recognition}} &
               \shortstack{{\bf Traffic}\\{\bf classification}} \\
      \hline               
      Dataset & FEMNIST & Huawei \\
      Availability &  Public & Proprietary \\
      Samples & 805k  & 1.9M \\
      Sources   & \shortstack{3.5k\\writers}  & \shortstack{2\\networks} \\
      Classes & 62    & 35\\
      \hline
    \end{tabular} 
    \end{center}
\end{table}

\begin{figure}[!t]
  \begin{center}
  \includegraphics[width=0.8\linewidth]{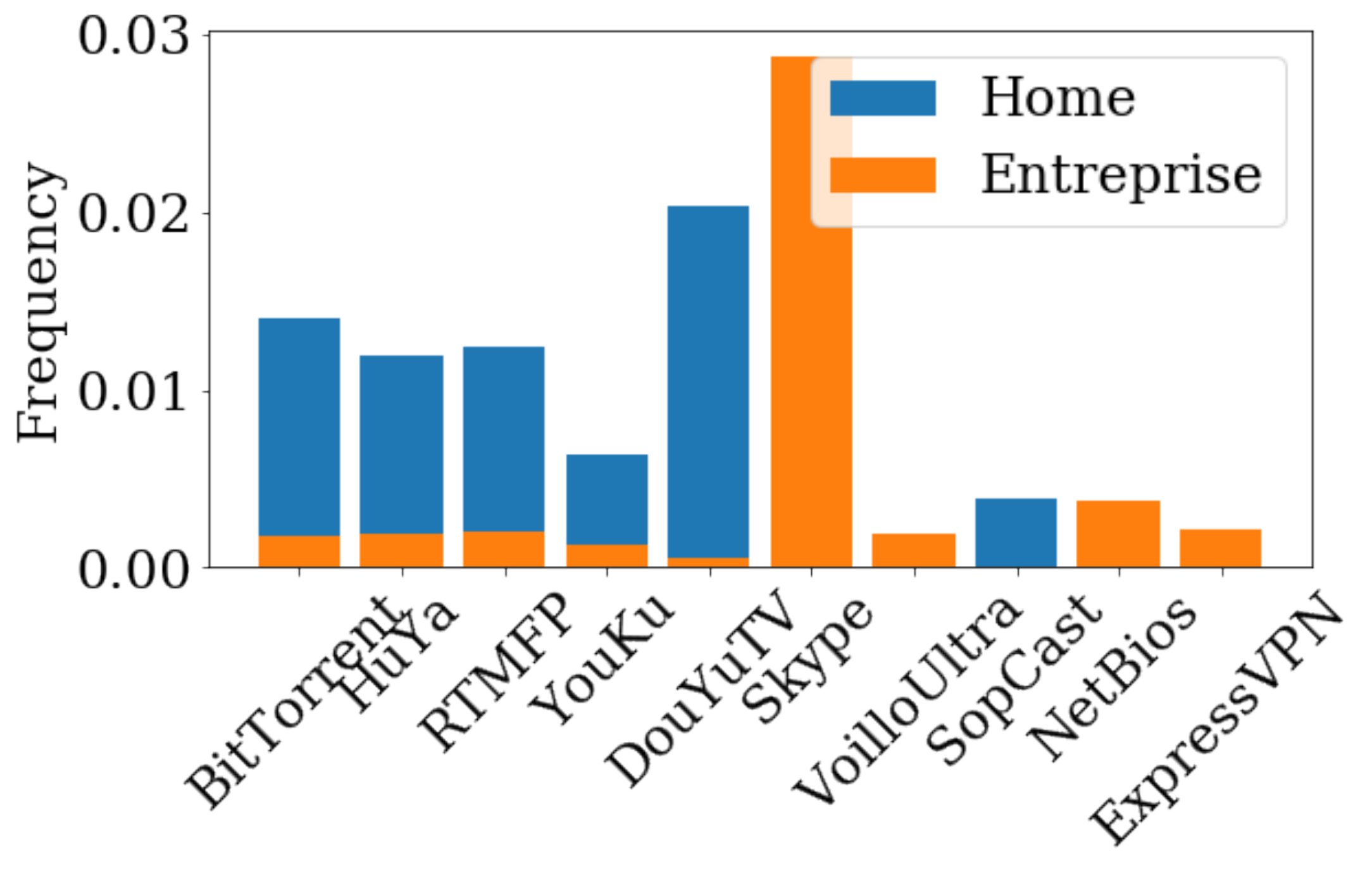}
  \includegraphics[width=\linewidth]{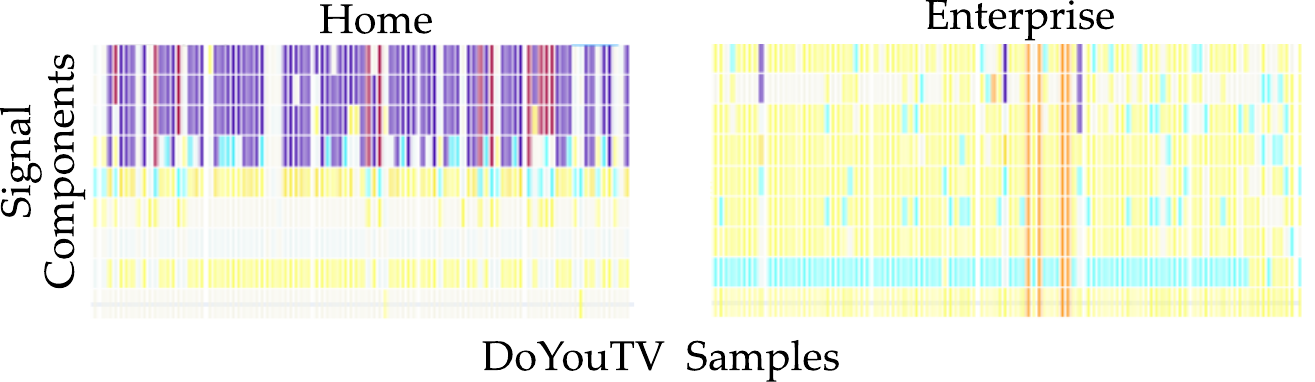}
  \caption{Example of real-world data characteristics in the Huawei TCP/IP dataset: non i.i.d and disjoint data distribution for 10 classes (top plot); example of multi-modal signal for the same class across two clients (bottom plot).
  }
  \label{fig:dataset}
    \end{center}
\end{figure}

Network traffic comes from two different environment: namely home users' households vs enterprise business, that can be considered as clients in the FL terminology.  This client mixture explains non i.i.d properties of the data: the top plot of Fig.\ref{fig:dataset} depicts 10 samples showing that, as expected, some applications are prevalent in the enterprise environment  (such as Skype for business, ExpressVPN,  NetBIOS, etc.) while others are exclusively present in the home environment (such as SopCast) so that class labels are even \emph{disjoint} in the two networks.  

Further, from Fig.\ref{fig:dataset}-(bottom) notice that applications  present in both datasets (DouYouTV in the example) can further exhibit signal multi-modality, as clearly shown in the heatmap of the signal component (y-axis, color encodes the size and direction of the first 10 packets from bottom to top) for different traffic samples (x-axis) in the Home vs Enterprise clients. 
Multi-modality is related to the physics of the underlying network and its interaction with specific applications: e.g., for some applications,   different environments  tied to specific network technologies (e.g., access type, encapsulations) and configuration (e.g., firewalls, NATs)  trigger different behavioral modes, which yield to specific and non portable signatures.
While this is perfectly normal and not done on purpose (e.g., to try to ``evade'' traffic classification), the resulting effect on training can be seen as adversarial with respect to what discussed in~\cite{Sattler2019, Zhao2018}, and need to be dealt with.

\section{Method}
Our proposed approach relies on convolutional neural networks architectures. CNNs are commonly composed of first stacked convolutional layers and then fully connected layers.  \cite{Yosinski14} demonstrated that, moving  from first layers to last layers, the features also transit from generic to specific. The network parameters  can be categorized into two groups:  generic parameters are common across different tasks, whilst task specific parameters convert it to high level representation.

\begin{algorithm}
\caption{HDAFL}\label{algo:hdafl}
\begin{algorithmic}[H]
\STATE \textbf{Server executes}:
\STATE  initialize $wb_0, wl_0$ 
\FOR {$each client k in parallel$}
\STATE $ClientInit(k, wb_0, wl_0)$
\ENDFOR
\FOR {each round t=1,2,\ldots,T}
  \STATE $S_t$=(random set of C clients)
  \FOR{each client k in parallel:}
  \STATE $wb_{t}^{k} \leftarrow$ ClientUpdate(k, $w_t$)
  \ENDFOR
  \STATE $wb_{t+1} \leftarrow \sum_{t=1}^{K} \frac{n_k}{\sum_{k=1}^{K}n_k} wb_{t}^{k}$
  
\ENDFOR
\STATE $\quad$
\STATE \textbf{ClientUpdate}(k, $wb_{t}$)
\STATE $wb_{t}^k \leftarrow wb_{t}$
 \FOR{each local epoch e=1,2,\ldots,E}
 \STATE Update $ wb_{t}^k, wl_{t}^k $ on local data
 \ENDFOR
 \STATE return $wb^k_{t}$ to server
\STATE $\quad$ 
\STATE \textbf{ClientInit}(k, $wb_0$, $wl_0$)
\STATE $wb_{0}^k \leftarrow wb_0$
\STATE $wl_{0}^k \leftarrow wl_0$
\label{algo:hdafl}
\end{algorithmic}
\addtocounter{algorithm}{-1}
\end{algorithm}

Following this split of generic and specific layers, we propose to share the generic feature extraction, like the convolutional layers, between the servers and the clients and more precisely to keep the specific layer as the classification layer only on the clients side.
This scheme allows the feature extraction part to benefit from all the data and therefore to increase its robustness. Each client can also adapt more to the local data distribution with the specific layer.
Additionally, keeping layers specific to the clients decrease the amount of bytes exchange between the clients and the server, in comparison of sharing the whole model. 

Our proposed Heteregoneous Data Adaptive Federated Learning (HDAFL) algorithm is illustrated as Algorithm~1.  In terms of notation, we denote with $wb$ the generic parameters and with $wl$ the specific local layer parameters: $wb_{t}^{k}$ indicates the model for client $k$ out of the total number of clients $K$ at $t$-th round of communications, and $n_k$ denotes the total number of samples on client $k$. The two free parameters $C$  and $T$ indicate the number of clients sampled at each round, and the maximum number of rounds respectively. Within each round, the local learning process at each client is governed by the number of epochs $E$.

The whole process starts with an initialization step: the server initializes the full parameters $wb_0, wl_0$ by using common initialization methods (such as random initialization, that draws value from normal Gaussian distribution). The server then broadcasts the initialized parameters to all the clients. 
At each round $t$ of communications, the clients update the local parameters by copying the generic parameters $wb_t$. For the first round of communication, the specific parameters are copied as well. The clients update the received model on their local data with fixed epochs $E$ and send back the updated generic parameters  $wb^k_t$ to server. 
Upon receiving the updates from the selected clients,
the server aggregates the generic parameters $wb^k_t$ by using averaging weighted by the number of total samples $n_k$ at each clients. Another other option is weighted by the norm of gradient, that we do not consider in what follows for lack of space.
The server then broadcasts the aggregated generic
parameters to clients. Stop conditions can include attaining
a minimum accuracy target, or a maximum number
of rounds $T$ (that we limitedly consider in what follows
without loss of generality).

\section{Experimental results}
We start by showing results on an image recognition use-case with the FEMNIST dataset, 
to contrast general properties of our federated learning approach on a repeatable dataset,
and next briefly assess its performance on the TCP/IP networks traffic classification use-case.

\subsection{Image recognition}

\subsubsection{Experimental settings}
We design three experiments with the FEMNIST dataset according to the distribution of the data on each client.
We consider the data related to 2000 writers which is a subpart of the whole FEMNIST dataset.
We implement three kinds of data sampling methods to distribute the data into clients: i.i.d.,  non i.i.d. and disjoint. The two latter cases reflect  real life scenarios, where each client has its own local data, whose distribution is not identical between clients. 

\begin{figure}[!t]
  \begin{center}
  \includegraphics[width=0.48\columnwidth]{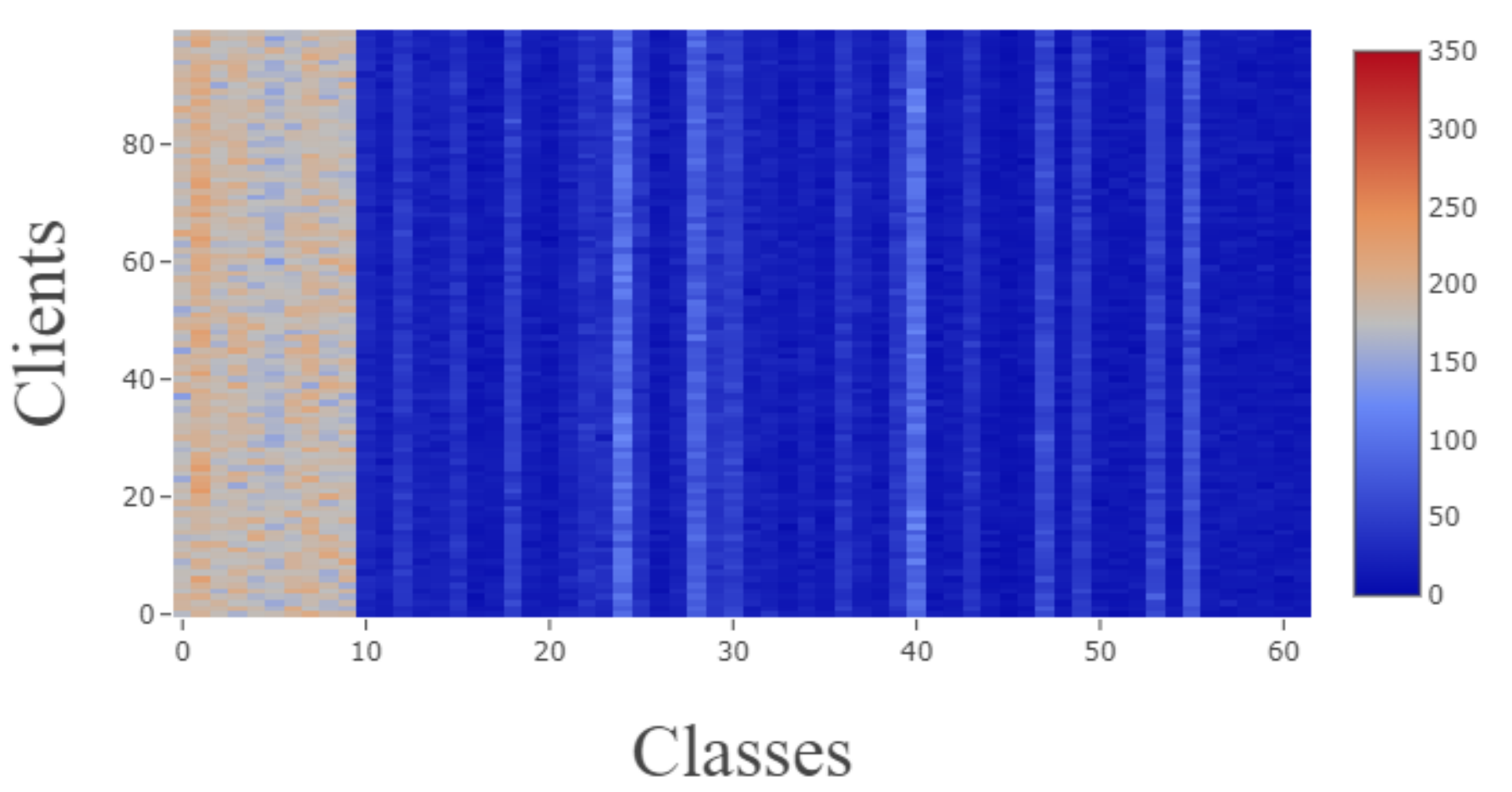}
  \includegraphics[width=0.48\columnwidth]{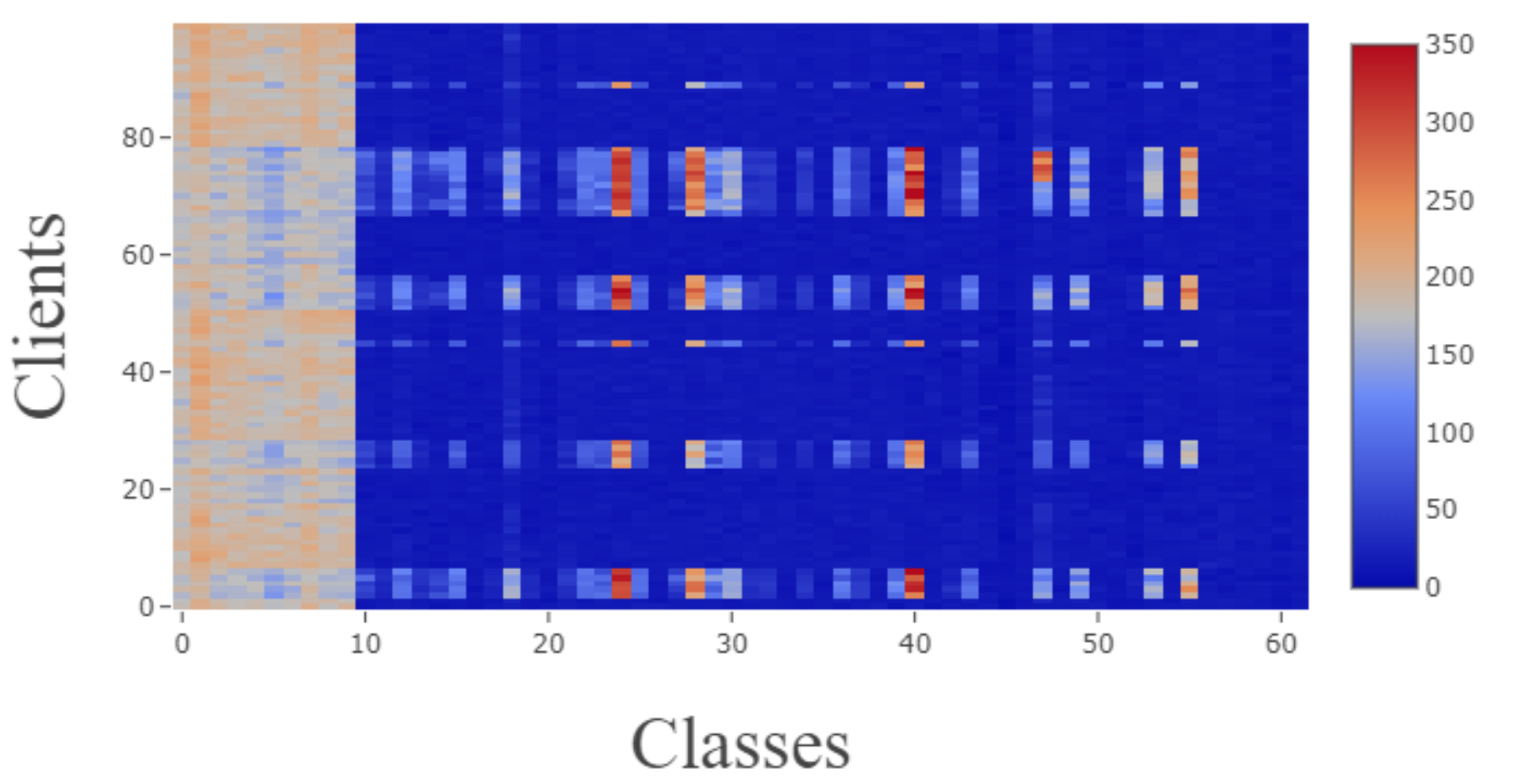}
  \caption{Client data distribution heatmap: (left plot)  i.i.d data distribution;  (right plot) non i.i.d data}
  \label{fig:heatmap}
    \end{center}
\end{figure}

The \emph{i.i.d sampling policy} imposes same data distribution on each clients. We mixed all the data and redistribute the data into $K=100$ clients with random selection. Thus, i.i.d data distribution is artificially obtained by redistributing data, breaking natural data imbalance: while this makes algorithm evaluation convenient, it also hides potential convergence problems.
For the sake of illustration,  the top plot of Fig.~\ref{fig:heatmap} reports a heatmap of the class per client distribution under i.i.d sampling.

The \emph{non-i.i.d sampling policy}  is done by considering writer of the FEMNIST dataset. The data are split on 100 clients and a client local data corresponds to several writers. 
The non-i.i.d sampling preserve the natural imbalance in the raw data distribution by considering each writer as a client, each of which has thus different data distributions.
The bottom plot of Fig.~\ref{fig:heatmap} clearly shows the imbalance in the non-i.i.d. case.

Finally, the \emph{disjoint sampling policy} is obtained by mixing all the dataset and attributing two classes per client. Comparing with highly non i.i.d sampling method used in \cite{Liang2019}, this sampling ensure there are no overlapping classes presented in two clients.
Thus, the data is artificially redistributed to extremely exacerbate imbalance on purpose. Notice that, given that the total number of classes is 62, this limit the maximum number of clients to 31 in this scenario.   

The CNN architecture is composed of two convolutional layers and two fully connected layers. We define the two convolutional layers and the following fully connected layer as the generic parameters, and the last layer as the specific parameters.

At each round of local training, $C$ clients are selected randomly among the available clients. We conducted test with different settings of $C\in\{10,20,50\}$ for the i.i.d and non-i.i.d settings and $C\in\{10,20,31\}$ for the disjoint settings (since we are capped by the number of classes in the latter case).
In order to have a fair comparison between algorithms, the clients chosen at each round are the same for a defined setting (number of client, client local data distribution).

The experiments are repeated 6 times with local epoch $E$=1 and a maximum of $T$=400 rounds of communications. We haven't conducted exhaustive search for optimum hyperparameters like batch size, learning rate, etc. 

\subsubsection{Accuracy performance}
Tab.~\ref{tab:acc_iid} summarizes the results for the three sampling strategies. In the i.i.d case, FedAvg by \cite{McMahanMRA16} gives better results than our approach.
In a centralized learning, having the whole dataset instead of a sub-part will improve the accuracy.
As the local data distribution is the same on each client, what would be beneficial in a centralized scheme is also beneficial on each client. Intuitively, sharing the whole model allows to somehow exploit the whole dataset, and therefore includes more variability during the training that increases the model representation capabilities.
In our approach, even if the feature extraction also benefit from seeing the whole dataset, the specific layers relies on a smaller amount of data, leading to an accuracy decrease.

In the non-i.i.d case, the results are reversed, our approach gives better results than the FedAvg. In HDAFL, the client specific layer allows to capture more the client data distribution than the FedAvg approach.  As non-i.i.d. data is prevalent in the real-world over  i.i.d. data, we also expect gains over FedAvg to be consistent in practice.

The disjoint experiment is an extreme case, that illustrates a limitation of the FedAvg approach, and project the potential benefit of our proposal.
To understand why this happens, consider that \cite{Zhao2018}  show that this disjoint case correspond to the maximum probability distance, thus maximum weights divergence. Consider further that in the disjoint settings, each client has two classes. For example, averaging the classification layer over 10 clients leads to add 9 times a non relevant weight (in the sense the class has not been seen during the training) and 1 time a relevant weight for triggering the neurons corresponding to each class. Our approach do not suffer from this problem, since it keeps the classification layer specific to each client. 

One potential downside of HDAFL is that last layer specificity might leads to some over adaptation.  However, we point out this may happen in cases where most classes are disjoint, which is an extreme case considered here only to show the potential gain of the method even where the benefit from joint learning is marginal. At the same time,  we expect real world cases to exhibit mixture of non-i.i.d. and occasional disjoint classes: in these cases, sharing generic model parameters is beneficial and client over adaptation less likely since several clients have their own data for the very same classes.

\rowcolors{3}{}{white}
\begin{table}[t]
  \begin{center}
    \caption{Accuracy results on FEMNIST dataset at  400 rounds of communications for the three experimental settings.}
    \begin{tabular}{llccc}
      \rowcolor{lightgray}    
      \multicolumn{2}{c}{Experimental}     & \multicolumn{3}{c}{Client selected}  \\
      \rowcolor{lightgray}
      Settings & Method &  10 &  20 &  50   \\
      \hline
      i.i.d & HDAFL   & 74.90 & 75.37 & 75.89\\
       &    FedAvg    &  \textbf{78.26} & \textbf{79.00} & \textbf{79.03}\\
      \hline
      non-i.i.d & HDAFL & \textbf{79.61} & \textbf{79.97} & \textbf{80.47}\\
            & FedAvg  &78.00 & 78.88 & 79.06\\
      \hline
      \rowcolor{lightgray}
      Settings & Method &  10 &  20 &  31\\
      \hline
      Disjoint & HDAFL &\textbf{97.53} & \textbf{97.80} & \textbf{97.78} \\
            & FedAvg        &17.44 & 18.99 & 19.51\\
      \hline

    \end{tabular}
\label{tab:acc_iid}
\end{center}
\end{table}


\subsubsection{Communication complexity}

\begin{figure}[t]
  \centering
  \subfloat[10 Clients]{\includegraphics[width=.5\linewidth]{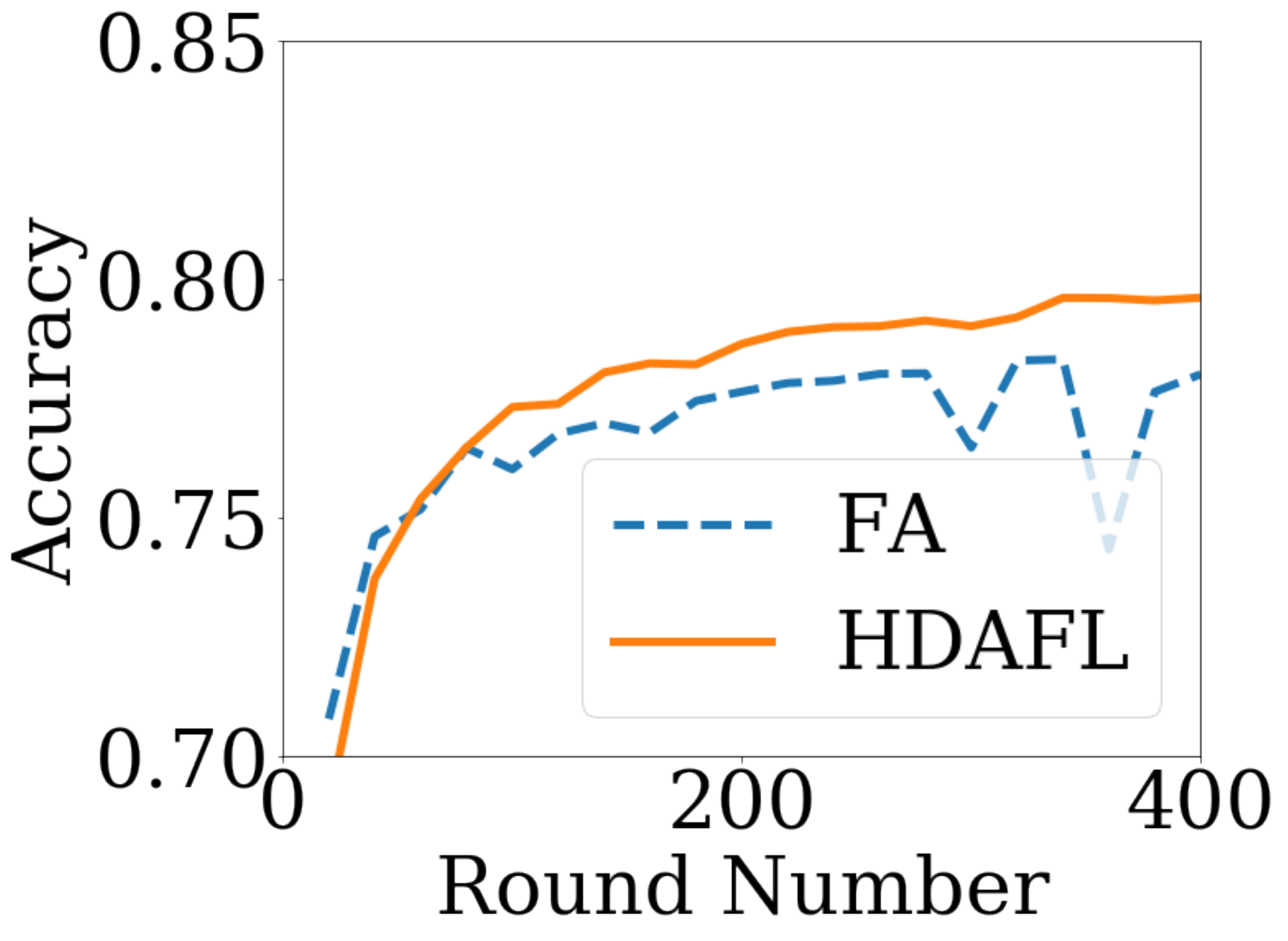}}
\subfloat[50 Clients]{\includegraphics[width=.5\linewidth]{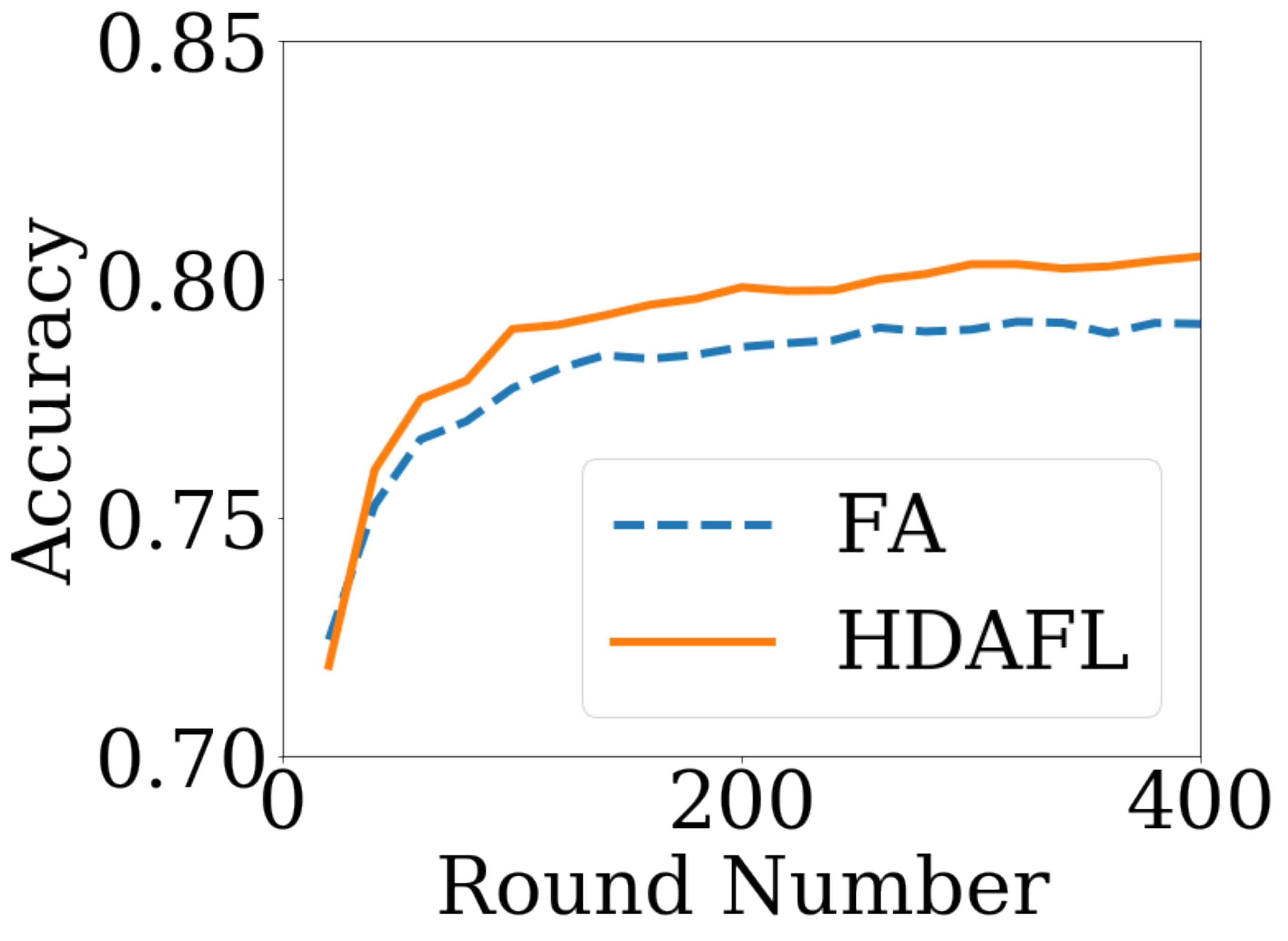}}

\subfloat[10 Clients]{\includegraphics[width=.5\linewidth]{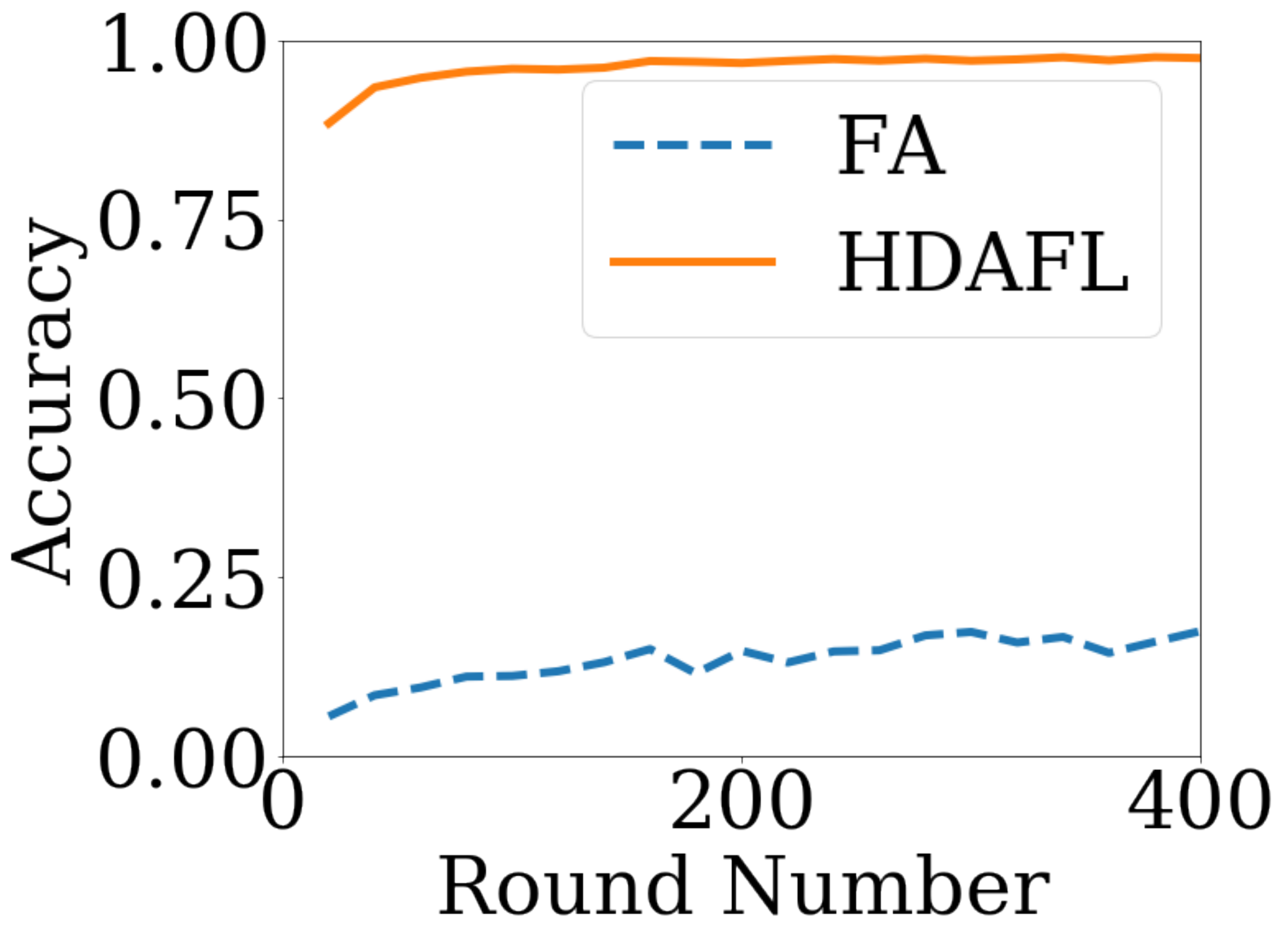}}
\subfloat[31 Clients]{\includegraphics[width=.5\linewidth]{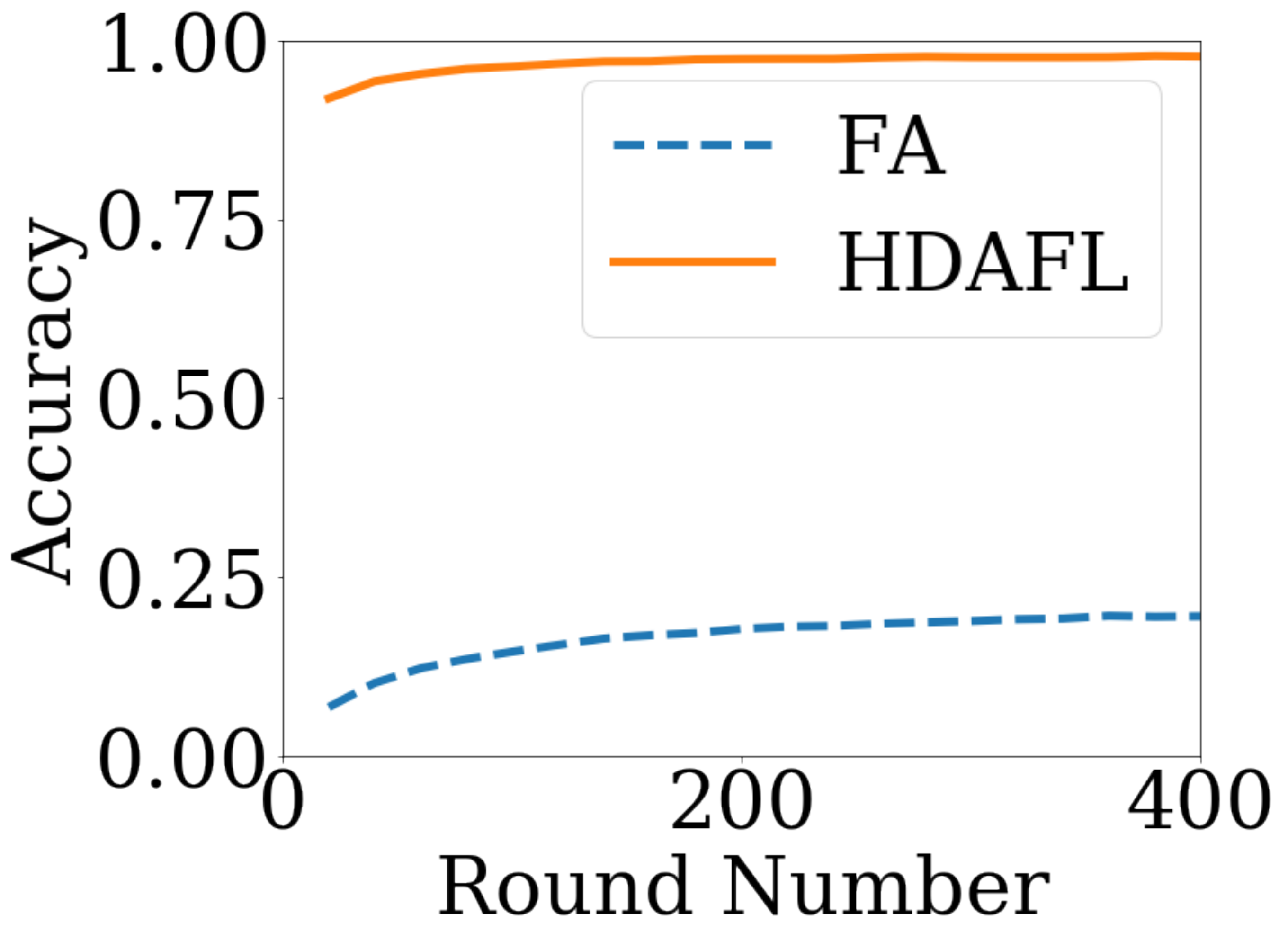}}
\caption{Clients average testing accuracy evolution over the 400 rounds of communications: non-i.i.d (top a,b plots) and disjoint (bottom c,d plots) sampling policies.
}
\label{fig:convergence}
\end{figure}

Communication cost is an important metric for federated learning. Two factors concur in reducing communication complexity:  the rate of convergence at a target accuracy,
which determines the number of rounds, and the amount of model parameters transferred at each round.
We show in Fig.~\ref{fig:convergence} the convergence analysis of the accuracy over the number of rounds communications for the non-i.i.d sampling.
Our method converge more quickly than the FedAvg and achieves a higher accuracy. Generally speaking, curves are smoother when more clients participate in learning. This smoothness can also be attained by our method with less clients: sharing only the generic parameters decreases the weights divergence, which in turns increases the model stability, explaining the smoothness. Reduced weight divergence leads to faster convergence and smaller communication costs. 

Tab.~\ref{tab:communication_cost} report the communication cost needed to attain a target of 78\% average accuracy\footnote{To allow a full comparison, target is selected as the minimum over non-i.i.d settings (FedAvg $C=10$ in Tab.\ref{tab:acc_iid})}. The minimum communication cost of HDAFL is 86.8 Mbyte, two times lower than FedAvg. If training duration is a priority factor, we can let 50 clients  participate in each round, so that learning lasts only 83 rounds, which is 1.3$\times$ faster than FedAvg and requires a  1.5$\times$ smaller volume of exchanges.  We stress that these already quite sizeable gains are obtained in the non-i.i.d case, and can be expected to grow when the imbalance grows, as testified by the  performance gap in the extreme disjoint case.

\rowcolors{3}{}{white}
\begin{table}[t]
  \begin{center}
    \caption{Communication cost to attain 78\% accuracy (non-i.i.d)}
    \begin{tabular}{lcccc}
  \rowcolor{lightgray}
  
  \shortstack{Clients \\selected} & Method  & \shortstack{Rounds \\number} & \shortstack{Communication \\Cost (MB)} \\
  \hline
  10 & HDAFL   & {\bf 139} & {\bf 86.8  } \\
  20 & HDAFL   & {\bf 110} & {\bf 137.5 }\\
  50 & HDAFL   & {\bf 83} & {\bf 259.3 }\\
  \hline
  10 & FedAvg  & 259 & 171.7\\
  20 & FedAvg  & 157 & 208.2\\
  50 & FedAvg  & 115 & 381.2\\
  \hline

\end{tabular}
\label{tab:communication_cost}
\end{center}
\end{table}

\subsection{TCP/IP Traffic classification}

We note that in the TCP/IP Traffic classification use case, FedAvg could be hardly used in practice due to the privacy constraint that HDAFL solves. Still, it makes sense to contrast the performance of both algorithms.
The algorithmic targets are to contain the model accuracy loss due to distributed training to less than 1\% (hard constraint) with an at least 10-fold reduction 
in the communication cost with respect to the case where all data is sent in a 
space efficient and compressed format to a central location for training (soft constraint). 

As our focus is on the ability of FL algorithms to distributed the training process, we report results relative to the centralized model. As shown in Tab.\ref{tab:acc_tc}, our method always achieves less accuracy drop compared with the FedAvg\cite{McMahanMRA16}. In particular,   accuracy reduction in HDAFL approaches the 1\% target already after 4 rounds, with a 33-fold reduction of the communication cost. At 8 rounds, the accuracy reduction is half of the target, for a 16-fold communication cost.  In contrast, FedAvg is not able to achieve the accuracy target, and this irrespective of the number of rounds. Finally, it is once more worth stressing the slight but noticeable communication cost reduction of HDAFL over FedAvg, which stems from the confidentiality of the last layers.


\begin{table}[t!]
  \begin{center}
    \caption{Comparison table for the TCP/IP traffic classification. Performance loss (accuracy) and gain (communication) are relative to those of a centralized trained model.}
    \label{tab:acc_tc}
    \begin{tabular}{lrrr}
      \rowcolor{lightgray}
       &   & Accuracy  &  Communication \\
      \rowcolor{lightgray}
      Method & Rounds & loss  &  gain\\
      \hline
             &4  & -2.0\% &  28.9$\times$ \\
      FedAvg &8  & -1.6\% & 14.5$\times$\\
             &20 & -1.7\% &  5.8$\times$\\
      \hline
            & 4 &  {\bf -1.2\%} & {\bf 32.9$\times$}\\
      HDAFL & 8 &  {\bf -0.4\%} & {\bf 16.5$\times$}\\
            & 20&  {\bf -0.5\%} &  {\bf 6.6$\times$}\\
      \hline
    \end{tabular}
\end{center}
\end{table}

\section{Conclusion}
Towards the deployment of federated learning in large scale production system,
we present our method that is specially designed for limiting the share of  sensitive  model information, and solving the practical problems non i.i.d data, disjoint classes and signal multi-modality across dataset.

The main idea behind our proposal is to share generic feature extraction among clients, while letting each client keep a local private version of the specific classifier.   Experiments show that, with respect to FedAvg by \cite{McMahanMRA16}, our method only exhibit a slight penalty in case of i.i.d data, while performance benefits (in terms of accuracy, convergence rate and communication cost) are noticeable for non-i.i.d. data, and significantly 
increase as the imbalance grows.

\newpage
\bibliographystyle{named}
\bibliography{ijcai20}

\end{document}